\documentclass{article}

\usepackage[round]{natbib}
\usepackage{url}  % Formatting web addresses  
\usepackage[latin1]{inputenc} %UNIX support if unicode package fails

\newcommand{\con}[1]{\texttt{#1}}
\newtheorem{defn}{Definition}

\title{An evolutionary approach to Function}
\author{Phillip Lord\\
School of Computing Science\\ 
Newcastle University\\
Newcastle-Upon-Tyne\\ 
United Kingdom\\ 
NE1 7RU\\
\url{phillip.lord@newcastle.ac.uk}}

\begin{document}

\maketitle

\begin{abstract}
  \textbf{Background:} Understanding the distinction between function and
  role is vexing and difficult. While it appears to be useful, in practice
  this distinction is hard to apply, particularly within biology.

  \textbf{Results:} I take an evolutionary approach, considering a series
  of examples, to develop and generate definitions for these concepts. I test
  them in practice against the Ontology for Biomedical Investigations (OBI).
  Finally, I give an axiomatisation and discuss methods for applying these
  definitions in practice. 
  
  \textbf{Conclusions:} The definitions in this paper are applicable,
  formalizing current practice. As such, they make a significant contribution
  to the use of these concepts within biomedical ontologies. 
\end{abstract}

\section{Background}
\label{sec:introduction}

Large parts of modern biology are aimed at answering questions about function.
For example, much of the Gene Ontology deals with molecular
function\citep{Ashburner2000}. In dealing with the social aspects of science,
roles are similarly important. It is clear, therefore, that function and role
are important concepts in biomedical ontologies and are prime candidates for
inclusion in an upper ontology. A coherent, consistent and shared definition
for function and role is likely to decrease the effort required to integrate
independently-developed ontologies.

One upper ontology, the Basic Formal Ontology (BFO)\citep{bfo}, currently in
use by a number of groups, includes a definition of function. However, this
definition is not naturally applicable to biology; it is not clear, for
instance, that Gene Ontology molecular functions are also BFO functions, as
discussed later (Section~\ref{sec:biological-function}). An alternative
is available within the General Formal Ontology (GFO), which provides an
ontology of function\citep{Burek2006}. This ontology provides a more extensive
framework for describing functions but, in itself, does not define biological
function. Despite this, there is a reasonable degree of informal agreement
among biologists as to the meaning of the word. Conversely, while formal and
informal definitions for role seems clear, many people have difficulties in
applying it in practice.

In this paper, I address two key issues relating to the modeling of function
and role coherently for biomedical ontologies: firstly, how I unify the
definitions of function as they apply to artifacts and to life; and, secondly,
how do I differentiate between roles and functions? I consider illustrative
examples where the answers are reasonably clear and evolve a definition from
them. I then consider examples of the application of current definitions from
BFO in a practical biomedical use. Finally, I offer an axiomatisation in OWL
stemming from my definitions and consider how they could be applied in
practice.

\section{Results}

\subsection{Biological Function}
\label{sec:biological-function}

First, consider the current definition of function provided by BFO (see
Def:~\ref{df:function}). 

\begin{defn}
\label{df:function}
Function is a realizable entity the manifestation of which is an essentially
end-directed activity of a continuant entity being a specific kind of entity
in the kind or kinds of context that it is made for
\end{defn}

As a simple example, a hammer (\emph{the continuant entity}) was made to
hammer nails (\emph{the function}) in a hammering process (\emph{the
  end-directed activity}). This definition is problematic for biological
systems. The problem here is simple: most biological systems were not made or
designed for any purpose.

Although, it has not been incorporated into BFO yet, there is a potential
definition for ``biological function'' which would become a new child of
function (Def:~\ref{df:bfo_bio_function})\citep{Arp2008}. 

\begin{defn}
  \label{df:bfo_bio_function}
A biological function is a function which inheres in an independent continuant
that is i) part of an organism and ii) exists and has the physical structure
as a result of the coordinated expression of that organism's structural genes. 
\end{defn}

As an example, a foot (\emph{the independent continuant}) is part of an
organism, exists, and develops in a controlled way as a result of gene
expression. There are some difficulties with this definition. Consider the
following examples: a differentiated tumour has its structure through
coordinated expression of its genes, it exists and it engages in end-directed
activity; a male ant has its structure as a result of gene expression, and
engages in end-directed activity, however, it is not part of an organism. From
this, I conclude that a differentiated tumour does, indeed, have a function
(growing). A male ant, however does not have a function. Also consider
molecular function: the physical structure of a protein is independent of the
expression of an organisms structural genes -- only its presence depends on
this. A protein, therefore, does not have a function, by this definition. So,
there are two key problems: the definition does not work for entities above or
below a certain size; and most biological entities have their structure as a
result of coordinated expression. I offer the following alternate definition
(Def:~\ref{df:hom_bio_function}).

\begin{defn}
  \label{df:hom_bio_function}
  A biological function is a realizable entity that inheres in a continuant
  which is realized in an activity, and where the homologous structure(s) of
  individuals of closely related and the same species bear this same
  biological function.
\end{defn}

The definition given uses the notion of homology; evolution is key to our
understanding of biology and it is appropriate that it should be used to
define biological function. If a biological structure has a function, then
this function will have evolved along with the structure; so, other structures
with a common evolutionary descent will display the same behaviour. This
definition also mirrors closely normal biological practice; the most common
way to determine the function of an unknown structure is to look for function
of a homologous structure.

It should be noted that this definition of biological function is \emph{not}
circular, although it has itself as part of its definition; rather it is
recursive; a chimp hand and a human hand can have the same function because of
each other. It does require that a structure must have a homolog for it to
have a function. It does not require that these homologous structures be
extant.

Applying this to our examples: the tumour now has no function because it has
no homologs (different tumours arise as independent events and share no common
ancestor). Likewise, the activity of the male ant now clearly is a function,
as many different, related organisms behave in a similar way. Finally, a
protein may have a function depending on the activity of its homologs. In
short, this definition results in the same conclusions as our biological
understanding.

\subsection{Relating the Functions}
\label{sec:relating-functions}

As well as biological function, another subclass, \textit{artifactual
  function} has been suggested\citep{Arp2008}. Next, I consider the
relationship between function and these biological and artifactual subclasses.
Taking an illustrative example, consider the sole of my foot and the sole of
my shoe. They appear to operate to: provide a frictional surface to enable
motion; provide padding to \textit{reduce shock} to everything above; be tough
enough to resist abrasion. They would appear to have the same function;
indeed, like many artifacts, we would guess that the shoe owes much of its
design to mimicry of biology. It would seem, therefore, sensible that
instances of \textit{Shock Resistance} could be either a biological or
artifactual function. The alternative would be to duplicate many functions
under both subclasses (``biological shock resistance'', and ``artifactual
shock resistance'') with very similar definitions. Some functions such as
reproduction have to be a biological function, while others liquifying iron
can never be. In short, whether an instance of function is biological or
artifactual should be determined from the nature of the entity in which it
inheres, rather than the process by which it is realized. I therefore offer a
simple definition of function which reflects this (Def:~\ref{df:conj_func}).

\begin{defn}
  \label{df:conj_func}
  A function is a realizable entity which is a biological function or an
  artifactual function.
\end{defn}

For the purposes of this paper, I note that the definition given for function
earlier (Def:~\ref{df:function}), can serve as a reasonable definition for
artifactual function. 

It is also interesting that this definition covers some unusual but
non\-patho\-logical examples. Take a bacterium whose colonies change colour
depending on the presence of a toxin and which was produced using synthetic
biology techniques. The components have all evolved, but the organisation has
not. Is the detection of the toxin then a biological or artifactual function?
This is clearly a difficult, if uninspiring, question but given
Def:~\ref{df:conj_func}, one that can avoid by simply describing it as a
function; perhaps more intuitively, it can be described as both a biological
and an artifactual function, suggesting strongly that these two classes should
not be disjoint.

\subsection{Roles}
\label{sec:roles}

Next, having considered the definition of function and its applicability to
biology, I consider the issue of roles. The current definition
(Def:~\ref{df:bfo_role}) is complex; put more simply, it suggests that the
entity having that role can be involved in an activity but that it was not
necessarily intended for, nor necessarily has the structure for this.

\begin{defn}
  \label{df:bfo_role}
  A realizable entity the manifestation of which brings about some result or
  end that is not essential to a continuant in virtue of the kind of thing
  that it is but that can be served or participated in by that kind of
  continuant in some kinds of natural, social or institutional contexts. 
\end{defn}

Consider the relationship between role and function. Again, I shall use
a simple biological example, in this case of a man walking on his hands. By
our earlier definition (Def:~\ref{df:hom_bio_function}), ``to walk on hands''
is \emph{not} a function. While the homologous structure is, indeed, used for
walking on in all closely related species, most humans do not walk on their
hands. It would, therefore appear to be a role. In this context the hand has a
role of \textit{Shock Resistance}. This realizable entity also appears in the
hands of many other primates; in this case, however, it would appear to be a
function of the primate hand, as it is a function of their feet.

I am left with a similar conclusion as previously. Just as Shock Resistance
maybe either a biological or artifactual function, I must also conclude that
the individuals of the same class can be a role, depending on the nature of
the relationship between the independent continuant and realizable entity.

\subsection{OBI as a case study}
\label{sec:obi-as-case}

So far in this paper, I have considered a number of illustrative examples
and used these to draw conclusions about definitions for functions and roles.
This methodology is appropriate, but has the limitation that the choice of
other examples may have led us to different conclusions. In this section,
therefore, I will consider the use that OBI (Ontology for Biomedical
Investigations)\citep{soldatova2009} has made of function and
role (Analysis was performed on OBI rc-1, (release 2009-11-06)). I
choose to use OBI as it was built after BFO and with knowledge of it; many of
the ontologies available from OBO were started without its use or knowledge.

\begin{table}[]
  \begin{tabular}{| l | p{22em} |}
    \hline
    Function Bearer & OBI Function\\\hline
    Human & Perturb, Measure, 
    Separation (3), Sort \\\hline
    Computer & Information Processor (3),
    Consume Data\\\hline             
    Highly Generic & Freeze, Heat,
    Environment Control,
    Mechanical,
    Record, Contain,
    Transfer, Cool, Connection,
    Synthesizing, Excitation (2),
    Ionization, Energy Supply (1)\\\hline
    Distant Galaxy & Magnify\\\hline
    Out of Scope & Molecular Function (3)\\\hline
  \end{tabular}
  \caption{\small{OBI Functions considered as Roles. I provide suggestions for entities that
      could engage in the same end-directed activity, but without being designed for
      the purpose. \con{Function} has been omitted from the OBI term names. Numbers
      indicate child terms which have been omitted for brevity. I do not consider
      \con{Information} Processor to be the function of a computer, as the definition
      is more specific than the term suggests.}}
    \label{tab:functions_as_roles}
\end{table}

Considering first the functions of which OBI has 38. Can these functions be
fulfilled by an entity which was not designed for the purpose? As shown in
Table~\ref{tab:functions_as_roles}
most of them can, often by considering a
highly generic device (like a computer) or organism (like a human). Some are
highly generic in themselves and can be performed by many things (heat for
example). I find no cases where a function could not be fulfilled by an entity
which was not designed for the purpose. I consider ``Molecular Function'' to
be out-of-scope for this section as it is not clear whether it fulfils the BFO
definition of function.

Next, I consider OBI roles. There are many more roles than functions -- around
90 in fact. Due to the size, here I consider whole branches of the role
hierarchy.

There are a set of roles relating to reagents and their states. While label
role (defined as a reagent role realized in a detection of label assay) seems
sensible, it would appear that S$^{35}$ CTP, used to label DNA, is
manufactured specifically for this purpose. It certainly does not occur in
nature. It would appear to fulfil the requirements for a function.

Next, let us consider reference substance roles (a role which can support the
observation of similarities, differences, relative magnitude or change). In
many cases, biological assays use a reference which is not manufactured.
However, consider $\lambda$-HindIII fragments or a calibration standard. These
would all appear to function as an reference and have been produced
specifically for this purpose.

Finally, there are a number of roles for molecules or organisms: antigen role,
pathogen role and primer role. In an age where we can engineer the production
of DNA, protein and organisms, it is not clear that these can only ever be
roles and not functions. 

As a result of this analysis, I suggest a modification to the current
definition of role (Def:~\ref{df:bfo_role}). Both roles and functions can
become apparent (\emph{realized}) in natural, social or institutional
contexts. That such a context exists does not provide a clear differentation
between role and function; the critical distinction relates to whether the
entity in question was designed to be or has homologs that are engaged in a
given process. I suggest, therefore, this alternate and simpler definition for
role (Def:~\ref{df:simple_role}).

\begin{defn}
  \label{df:simple_role}
  A realizable entity the manifestation of which brings about some result or
  end that is not essential to a continuant because of its kind. 
\end{defn}

In short, from this case study, I conclude that the role/function distinction
is not clear. While OBI has a specific intent in mind with its application of
the distinction (broadly and not exhaustively, social or experimental roles,
device or instrument functions), this distinction is not the distinction made
in the current definitions of role and function in BFO; further given that
most functions could also appear to be roles, and many roles appear also to be
functions, I suggest that the distinction made in the current definitions is
not useful in the context of OBI. 

The earlier theoretical analysis seems to be confirmed in practice within OBI.
This suggests that the limitations in the definitions drawn from the earlier
illustrative examples are general and not simply as a result of the specific
examples chosen.

\subsection{An axiomatisation for function and role}
\label{sec:an-new-axiom}

I have produced an axiomatisation in OWL of functions and roles as defined in
this paper, available as described in the abstract; due to space considerations
I report here key differences between this and the BFO axiomatisation.

\begin{itemize}
\item There is an explicit relationship between \con{RealizableEntity} and
  \con{Process}. Subclasses use a more specific relationship. So, a
  \con{ToAbsorbShock} function may only be realized by a
  \con{AbsorbtionOfShock} process, if it is realized at all.
\item \con{Function} and \con{Role} are defined classes. Stating that an
  instance of \con{To\-Absorb\-Shock} \con{is\_function\_of} instance of
  \con{FootSole} implies, therefore, that the former is a function.
\item Most leaves of \con{Realizable\-Entity} are direct children of
  \con{Realizable\-Entity}, with a few exceptions (\con{ToReproduce} is a child
  of \con{Biological\-Function}).
\end{itemize}

The definitions could be extended further; for simplicity, I have not added
classes to differentiate between organisms and artifacts. These could be added
to automate the population of \con{BiologicalFunction} and \con{ArtifactualFunction}. 

At the current time, it remains an open question whether \con{Role},
\con{Function} and its children should be disjoint. The key example of the
function of a synthetic biological organism suggests that \con{Biological} and
\con{ArtifactualFunction} function should not be disjoint (as it appears to be
both), but I have no example which suggests whether \con{Role} and
\con{Function} should be disjoint. In axiomatisating the examples given, these
disjoint statements make no practical difference.

Many ontologies are built using the OBO format; while this has a slightly
weaker semantics than OWL it is possible to represent much of OWL in OBO
format\citep{golbreich2007}. The axiomatisation presented here can be
represented using OBO \textit{union} and \textit{intersection of} to describe
classes, which is usually translated as a definition. The universal link
between \con{RealizableEntity} and \con{Process} has no natural equivalent.
However, as this link has its own specific relationship which is restricted to
this use, problems caused by the lack of an inexact semantic equivalent are
likely to be relatively minor.

The axiomatisation presented here is related to that produced by others;
Dumontier\citep{dumontier2008} focuses more on roles, while
Burek et al.\citep{Burek2006} provides for a more complex representation, covering issues
such as preconditions. 

\subsection{Applying the Definitions in Practice}
\label{sec:applying-definitions}

Finally, I consider whether these definitions are \emph{applicable}; for a
given set of entities how do we decide whether we have a function (of either
subclass) or a role. 

The definition of an artifactual function easily allows its application:
first, we determine whether the entity in question is an organism or part of
one (which it should not be); second, we could ask whoever produced the entity
what it was designed for. Of course, the second may not always be possible, in
which case, we can guess from its design what its purpose is. In most cases,
these questions will provide a clear answer.

For biological function, the situation is less clear. Whether an entity is an
organism or part of one is, in practice, likely to be straightforward for
extant entities; otherwise, we can apply palentological techniques.
Likewise, identification of closely related species and homologous structures
is well known as it forms the basis of taxonomy. While developing an exact
definition for ``closely related'' is outside the scope of this paper, it is
possible.

The definition that I introduce for Function in this paper (Defn
~\ref{df:conj_func}) is conjunctive; it is either biological or artifactual.
Here I have given little evidence that these are the only kind of function.
Fundamentally, these two arise from very different mechanisms. There could be
other appropriate subclasses of function; the most obvious possibility would
be Chemical Function. However, artifacts are designed by humans who
understand, mimic and improve on biology by building tools. It is this mimicry
that we wish to reflect with a common definition joining biological and
artifactual function; this is not true for Chemical Function.

To determine whether something is a role, it is possible to make a
determination on the basis of whether the context is optional\citep{Arp2008}.
However, this optionality is a difficult criterion; firstly, all
RealizableEntity's are optional in the sense that they might never be realized
and, secondly, the optionality can depend on how specifically we define the
bearer. A hammer is not designed to hammer nails, as claimed earlier, it is
designed to hit things; a nail hammer is designed to hit nails, a toffee
hammer to hit toffee, a warhammer to hit anyone who irritates you. In
practice, a role can be considered to be a negative definition; if there is a
continuant and an end-directed activity that the continuant can be involved
in, and this involvement is known not to fulfil the definition of either
function, then we have a role.

In this paper, I have considered OBI and found that the distinction between
role and function is hard to apply; this is not true for all ontologies. For
example, consider the Gene Ontology. In many cases, the homology will be
considered as a standard part of the operating procedure\citep{goevidence} in
determining the function of a gene product; regardless, the evidence codes
would allow us to make the distinction. We can conclude, therefore, that when
the Gene Ontology is used to annotate a protein, this describes a biological
function rather than a role.

\subsection{Life is hard}
\label{sec:life-hard}

So far, I have considered a set of examples and how the definitions might be
applied, including examples from OBI which have not been preselected. However,
categorising life is hard; here, I consider some examples which present
difficulties for the definitions I have given and the implications of these
examples.

In the first example, which I term a \emph{drop out} species, consider a human
walking on their hands. Earlier (Section~\ref{sec:roles}), I have
suggested that this should be considered a role. Most of the primates do,
however, walk on their hands. However, given that the homologous structure of
closely related species use the structure for the same purpose, the definition
of function (Def:~\ref{df:hom_bio_function}) would appear to apply. It is for
this reason that the definition specifies that (most) individuals of the same
species must also demonstrate this behaviour (This definition differs slightly
from that given at Bio-Ontologies 2009). In short, in the absence of most
individuals in a species using a structure in a specific process, we should
not use consider this structure to have a function.

The second example, I term a \emph{drop in} species. Again, using a human
example, I use my larynx for vocalisation and talking. Most primates,
likewise, vocalise with their larynx; therefore, according to the given
definition, this is a function of the larynx. However, speech is considered
unique to humans, and therefore, their larynx; given that homologous
structures in closely related organisms do not bear this realizable entity,
which is part of the definition for biological function
(Def:~\ref{df:hom_bio_function}), I am forced to conclude that this is a role
of the larynx and not a function. 

In short, while sharing a realizable entity within a species is NOT sufficient
to allow the conclusion that this entity IS a function, NOT sharing a
realizable entity within a species is sufficient to conclude that this entity
is NOT a function. 

One solution to this difficulty is to state that where most individuals in a
single species use a structure within a given process, this alone is
sufficient to conclude that the structure has a function. Simply, most humans
talk with their larynx, therefore this would be a function. I counter this,
however, with the example that most humans use their fingers to operate their
mobile phones, so we would be forced to conclude that this would also be a
function. As this seems opposed to normal biological intuition and usage, I
conclude, the presence of most individuals in a species using a structure in a
specific process, is \emph{not} sufficient to conclude that this structure has
a function.

It is also possible that this difficulty could be resolved with greater
knowledge or changes in biology. Def:~\ref{df:hom_bio_function} does not
require species be extant; if a close, but extinct, relative of humans were
shown to speak with their larynx, or if humans speciated while maintaining
their speech, again, I would conclude that this represented a function.

While human speciation seems unlikely, it is much more relevant to other taxa.
Bacteria, in particular, evolve rapidly. There are many genes and proteins in
bacteria which are unique to a species, family or lineage\citep{Siew2004}. In
this case, the requirement for closely-related species seems to rule out the
presence of a function. Again, this seems opposed to normal biological
intuition and usage. I would counter this with two arguments. First, unlike
primates, our knowledge of the extant bacterial species is very limited. The
lack of knowledge of another speaking primate species is good evidence that no
such species exists; the lack of knowledge of a close relative for a given
bacterial species is not. Second, any definition which relies on a notion of a
species is only as good as the definition of species; for bacteria, there is
considerable debate about the utility of a species
classification\citep{Rossello-Mora2003}; my definition of function will need
to evolve along with our understanding of bacterial ecology and gene flow; it
may be necessary, as has been suggested with definitions for bacterial
species\citep{Rossello-Mora2003}, to have different definitions of
BiologicalFunction tailored to different parts of the taxonomy. 

The evolution of a definition of role and function for proteins is difficult.
At the level of the protein, I side with
Dumontier\citep{dumontier2008}, who suggests that the
role/function distinction may be redundant; broadly, proteins can do anything
their structure allows, and only do things their structure allows. The
definitions given in this paper have a consistent interpretation at the level
of the protein; this avoids the necessity of deciding at which level of
granularity to stop making the role/function distinction. We can make the
distinction at all levels if we choose, but we are not forced to do so, at
those levels of granularity where it is not useful.

It is important to note that arguing against a role/function distinction for
proteins is not to dismiss the experience of biologists in the analysis of
function assignment for genes. In this sense, the word ``function'' is being
used to describe an association between a protein and a process that a protein
molecule may be involved in; in short, the word ``function'', in this case,
can be considered to be a synonym for ``realizable entity''.

\section{Conclusions}
\label{sec:discussion}

Here, I have taken an evolutionary approach to function and role by
considering examples and using this to derive definitions which are as
consistent as possible with current use within biomedical sciences. These
definitions have been encoded in an axiomatisation which should enable the use
of these definitions in a machine-interpretable way.

The applicability of these definitions is a key advantage; the current
distinction being made between function and role is a hard one to understand
and apply. My definition distinguishes between the two based on the nature of
the relationship to the independent continuant in which they inhere. I suggest
that it is very hard to make the distinction at the class level; my study of
OBI shows that very few of the functions and roles clearly fall into one
category or another. For an individual continuant bearing a realizable entity,
this distinction appears to be much more straightforward.

I also provide a definition of biological function, something that is
currently lacking in BFO. I have paid close attention to current biological
usage; the definition is close to the process used to determine function.
Moreover, it is highly applicable; all parts of the definition are measurable.

The desire for an applicable and measurable definition is also the reason that
I have avoided a definition based on the outcome of selective pressure; this
is hard to test in most circumstances, requiring expensive evolutionary
studies, and impossible for extinct species. Selective pressure can also be
transient. Consider industrial melanism\citep{majerus1998}; should melanic
coloring be considered to gain its function during periods of pollution and
lose it in post-industrial periods? By way of analogy, should a spanner
measured in inches be considered to lose its function following metrication?
Serendipitously, it also avoids difficult questions about artificial
selection; we can state clearly that cows do not have a function of producing
beef, though this is the outcome of selection.

Importantly, my definition of biological function works across multiple levels
of granularity: from organisms and organism parts through to genes and
molecules; this is not true of previous definitions\citep{Arp2008}, which
cover only anatomy. It is, however, not clear how useful the role/function
distinction is for proteins and genes, as discussed earlier
(Section~\ref{sec:life-hard}); It is for this reason that I have used homology
rather than orthology as the basis for the definition, as the latter is
limited to the genetic scale, where the distinction is least useful.

Finally, my definitions also do not allow distinctions that may often be made
between different types of function. For example, most biologists would
consider motion the most important function of muscle, while heat
production a byproduct; or, for a more pathological example after
\citet{hoehndorf2009}, most biologists would
consider blood circulation to be a function of the heart, but ``making loud
thumping noises'' not to be. This is a concern which could be best addressed
by incorporating a degree of social ascription into the categorisation of
realizable entities within biology; although it is outside the scope of this
paper, this would provide a valuable and useful addition to the current
ontological practice. 

In summary, I believe that the definitions and axiomatisation given in this
paper make a significant contribution to the use of role and function in
biomedical ontologies. They should enable a consistent use of these classes,
because they consider current usage of the terms and the applicability of these
definitions. I seek not to change current use but to formalize it.

\section{Competing Interests}

The author has no competing interests. 

\section{Acknowledgements}
\label{sec:acknowledgements}

Thanks to Frank Gibson, Allyson Lister, James Malone, Helen Parkinson, Matt
Pocock and Robert Stevens for many useful discussions on the contents of this
paper. Thanks to the audience at Bio-Ontologies 2009 for the example of
rapidly-evolving bacteria.

To the memory of Mike Majerus who showed me how fundamentally strange
evolution can be.

\bibliographystyle{abbrvnat}
\bibliography{phil_lord_evolving_a_definition_for_function}

\begin{thebibliography}{12}
\providecommand{\natexlab}[1]{#1}
\providecommand{\url}[1]{\texttt{#1}}
\expandafter\ifx\csname urlstyle\endcsname\relax
  \providecommand{\doi}[1]{doi: #1}\else
  \providecommand{\doi}{doi: \begingroup \urlstyle{rm}\Url}\fi

\bibitem[bfo()]{bfo}
{Basic Formal Ontology}.
\newblock URL \url{http://www.ifomis.org/bfo}.

\bibitem[goe()]{goevidence}
{Guide to GO Evidence Codes}.
\newblock URL \url{http://www.geneontology.org/GO.evidence.shtml}.

\bibitem[Arp and Smith(2008)]{Arp2008}
R.~Arp and B.~Smith.
\newblock Function, role and disposition in basic formal ontology.
\newblock In \emph{The BIo-Ontologies Workshop (at ISMB 2008)}, 2008.

\bibitem[Ashburner et~al.(2000)Ashburner, Ball, Blake, Botstein, Butler,
  Cherry, Davis, Dolinski, Dwight, Eppig, et~al.]{Ashburner2000}
M.~Ashburner, C.~Ball, J.~Blake, D.~Botstein, H.~Butler, J.~Cherry, A.~Davis,
  K.~Dolinski, S.~Dwight, J.~Eppig, et~al.
\newblock {Gene ontology: tool for the unification of biology. The Gene
  Ontology Consortium.}
\newblock \emph{Nat Genet}, 25\penalty0 (1):\penalty0 25--9, 2000.

\bibitem[Burek et~al.(2006)Burek, Hoehndorf, Loebe, Visagie, Herre, and
  Kelso]{Burek2006}
P.~Burek, R.~Hoehndorf, F.~Loebe, J.~Visagie, H.~Herre, and J.~Kelso.
\newblock A top-level ontology of functions and its application in the open
  biomedical ontologies.
\newblock \emph{Bioinformatics}, 22\penalty0 (14):\penalty0 e66--e73, Jul 2006.
\newblock \doi{10.1093/bioinformatics/btl266}.
\newblock URL \url{http://dx.doi.org/10.1093/bioinformatics/btl266}.

\bibitem[Dumontier(2008)]{dumontier2008}
M.~Dumontier.
\newblock Situational modeling: Defining molecular roles in biochemical
  pathways and reactions.
\newblock In \emph{OWLED 2008}, 2008.

\bibitem[Golbreich et~al.(2007)Golbreich, Horridge, Horrocks, Motik, and
  Shearer]{golbreich2007}
C.~Golbreich, M.~Horridge, I.~Horrocks, B.~Motik, and R.~Shearer.
\newblock {OBO and OWL: Leveraging semantic web technologies for the life
  sciences}.
\newblock \emph{Lecture Notes in Computer Science}, 4825:\penalty0 169, 2007.

\bibitem[Hoehndorf et~al.(2009)Hoehndorf, Kelso, and Herre]{hoehndorf2009}
R.~Hoehndorf, J.~Kelso, and H.~Herre.
\newblock Contributions to the formal ontology of functions and dispositions:
  An application of non-monotonic reasoning.
\newblock In \emph{Bio-Ontologies 2009: Knowledge in Biology}, 2009.

\bibitem[Majerus(1998)]{majerus1998}
M.~Majerus.
\newblock \emph{Melanism: Evolution in Action}.
\newblock Oxford University Press, 1998.

\bibitem[Rossell\'o-Mora(2003)]{Rossello-Mora2003}
R.~Rossell\'o-Mora.
\newblock Opinion: the species problem, can we achieve a universal concept?
\newblock \emph{Syst Appl Microbiol}, 26\penalty0 (3):\penalty0 323--326, Sep
  2003.

\bibitem[Siew et~al.(2004)Siew, Azaria, and Fischer]{Siew2004}
N.~Siew, Y.~Azaria, and D.~Fischer.
\newblock The orfanage: an orfan database.
\newblock \emph{Nucleic Acids Res}, 32\penalty0 (Database issue):\penalty0
  D281--D283, Jan 2004.
\newblock \doi{10.1093/nar/gkh116}.
\newblock URL \url{http://dx.doi.org/10.1093/nar/gkh116}.

\bibitem[{The OBI consortium}(2009)]{soldatova2009}
{The OBI consortium}.
\newblock Modeling biomedical experimental processes with obi.
\newblock In \emph{Bio-Ontologies 2009: Knowledge in Biology}, 2009.

\end{thebibliography}

\end{document}